\documentclass[10pt,twocolumn,letterpaper]{article}

\usepackage[pagenumbers]{cvpr} %

\definecolor{cvprblue}{rgb}{0.21,0.49,0.74}
\usepackage[pagebackref,breaklinks,colorlinks,allcolors=cvprblue]{hyperref}

\def\paperID{*****} %
\def\confName{CVPR}
\def\confYear{2026}

\title{FrameDiffuser: G-Buffer-Conditioned Diffusion for Neural Forward Frame Rendering}

\author{Ole Beisswenger\\
{\tt\small ole.beisswenger@student.uni-tuebingen.de}
\and
Jan-Niklas Dihlmann\\
{\tt\small jan-niklas.dihlmann@uni-tuebingen.de}
\and
Hendrik Lensch\\
{\tt\small hendrik.lensch@uni-tuebingen.de}
}

\begin{document}
\twocolumn[{%
\maketitle
\vspace{-1em}
\begin{center}
    \includegraphics[width=\textwidth]{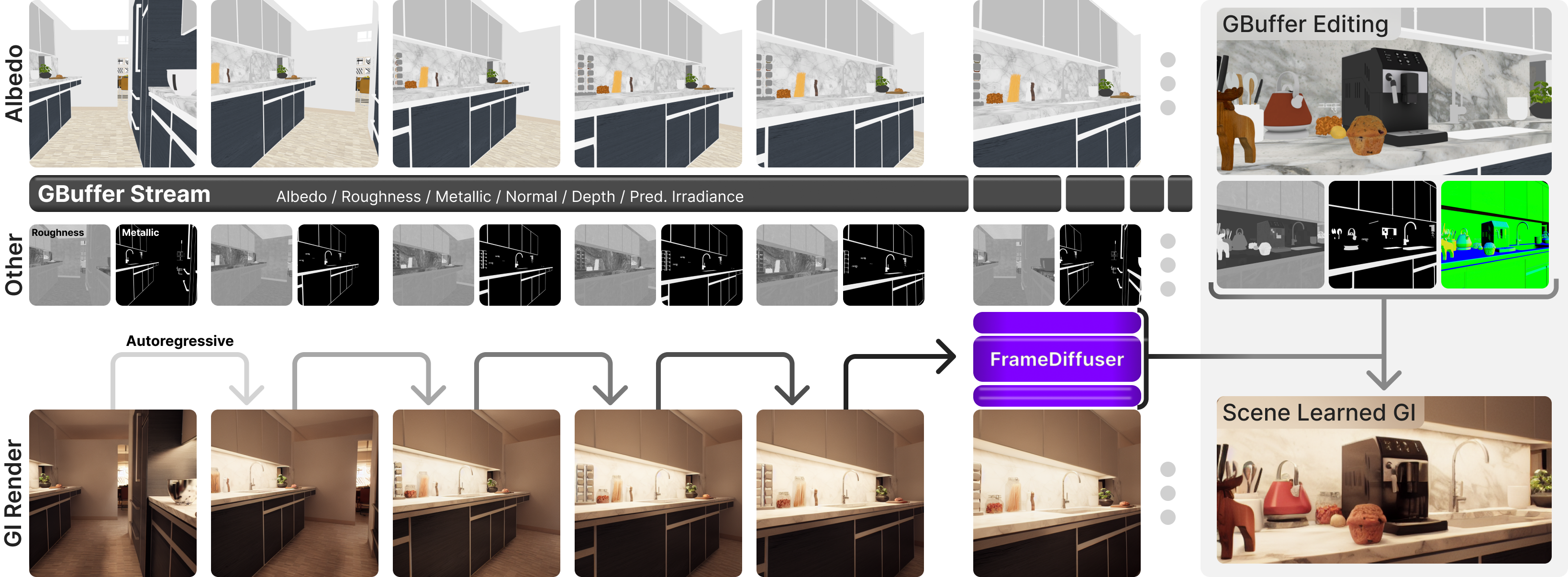}
    \captionof{figure}{\textbf{G-buffer to Photorealistic Rendering.} FrameDiffuser transforms geometric and material data from G-buffer into photorealistic rendered images with realistic global illumination (GI), shadows, and reflections. Our autoregressive approach maintains temporal consistency for long sequences, enabling neural rendering for interactive applications. Project page: \protect\url{https://framediffuser.jdihlmann.com/}}
    \label{fig:teaser}
\end{center}
\vspace{1em}
}]
\begin{abstract}
Neural rendering for interactive applications requires translating geometric and material properties (G-buffer) to photorealistic images with realistic lighting on a frame-by-frame basis. While recent diffusion-based approaches show promise for G-buffer-conditioned image synthesis, they face critical limitations: single-image models like RGB$\leftrightarrow$X generate frames independently without temporal consistency, while video models like DiffusionRenderer are too computationally expensive for most consumer gaming sets ups and require complete sequences upfront, making them unsuitable for interactive applications where future frames depend on user input. We introduce FrameDiffuser, an autoregressive neural rendering framework that generates temporally consistent, photorealistic frames by conditioning on G-buffer data and the model's own previous output. After an initial frame, FrameDiffuser operates purely on incoming G-buffer data, comprising geometry, materials, and surface properties, while using its previously generated frame for temporal guidance, maintaining stable, temporal consistent generation over hundreds to thousands of frames. Our dual-conditioning architecture combines ControlNet for structural guidance with ControlLoRA for temporal coherence. A three-stage training strategy enables stable autoregressive generation. We specialize our model to individual environments, prioritizing consistency and inference speed over broad generalization, demonstrating that environment-specific training achieves superior photorealistic quality with accurate lighting, shadows, and reflections compared to generalized approaches.
\end{abstract}
   
\section{Introduction}
\label{sec:intro}

Complete AI-driven simulation of interactive environments represents one of the most challenging objectives in neural computing. Projects like Oasis AI~\cite{oasisai2024} attempt to simulate entire games using a single neural network, generating each frame based on user input without any traditional game engine or rendering pipeline. However, even for visually simple games like Minecraft~\cite{oasisai2024}, these approaches struggle with fundamental issues: environments change dramatically when players turn around, geometry is hallucinated inconsistently, and control remains imprecise and laggy. 

Modern games employ deferred rendering pipelines that generate geometry buffers (G-buffer) storing per-pixel surface properties~\cite{saito1990,shishkovtsov2005}. Recent research has leveraged G-buffer for neural image and video synthesis~\cite{zeng2024,liang2025}.

RGB$\leftrightarrow$X~\cite{zeng2024} demonstrated that diffusion models can effectively leverage G-buffer information for high-quality deterministic single-image generation. DiffusionRenderer~\cite{liang2025} extended this concept to temporally consistent video sequences. These approaches leverage the precise geometric and material information in G-buffer as strong conditioning signals, reducing AI-hallucination or misinterpretations of the scene structure while still leaving room for neural models to generate the lighting and appearance in those bounds.

However, interactive applications such as video games are overlooked by current research in this field. These applications require frame-by-frame generation where each new frame depends on player input and therefore cannot be predetermined. Single-image models like RGB$\leftrightarrow$X~\cite{zeng2024} generate each image independently, lacking temporal consistency between consecutive frames. Video models like DiffusionRenderer~\cite{liang2025} need complete sequences at once, making them unsuitable for the sequential, frame-by-frame generation that interactive applications demand.

Furthermore, existing G-buffer-conditioned models aim for broad generalization across diverse environments. While valuable for general-purpose applications, real-world deployment in games often operates within specific visual domains. A game maintains its particular art style and lighting behavior throughout. Inspired by earlier work like EST-GAN~\cite{mittermueller2022}, which used environment-specific training with G-buffer conditions, we explore whether specializing models for specific domains could yield superior consistency compared to broadly generalized approaches.

In this work, we present FrameDiffuser, a neural rendering framework that fills the void through autoregressive, temporally consistent frame generation conditioned on G-buffer data. We train environment-specific models for six different Unreal Engine 5~\cite{unrealengine5} environments, demonstrating how specialization achieves superior consistency within specific domains. FrameDiffuser transforms G-buffer containing geometry and material properties into photorealistic rendered images with realistic lighting, shadows, and reflections. The model operates autoregressively using only incoming G-buffer data while conditioning on its own previously generated output, enabling stable generation over hundreds to thousands of frames. By combining precise geometric and material information from G-buffer with temporal conditioning on previous frames, our approach addresses the unique requirements of frame-by-frame generation for interactive applications. Our work makes the following contributions:

\begin{itemize}
\item \textbf{Dual-Conditioning Architecture}: We develop an architecture separating structural guidance (ControlNet with G-buffer + irradiance) from temporal coherence (ControlLoRA with previous frame latents).

\item \textbf{Three-Stage Training Strategy for Autoregressive Viability}: We demonstrate that our progressive training approach starting with black irradiance, introducing temporal conditioning, then self-conditioning is critical for preventing error accumulation in autoregressive generation.

\item \textbf{Irradiance-Based Temporal Lighting Guidance}: We introduce a novel approximate irradiance computation from previous frame outputs that provides strong temporal lighting cues while enabling autoregressive generation without ground-truth lighting information.

\item \textbf{Environment-Specific Specialization}: We demonstrate benefits of domain-specific training over generalization for interactive applications.
\end{itemize}

Our approach explores a path where AI augments and supports traditional rendering, preserving artist control over world building while demonstrating potential for individualized AI-based augmentation in interactive applications. By working within the constraints of frame-by-frame generation and embracing domain specialization, we investigate how neural rendering can complement existing pipelines rather than attempting to replace them entirely.

\begin{figure*}[t]
    \centering
    \includegraphics[width=\textwidth]{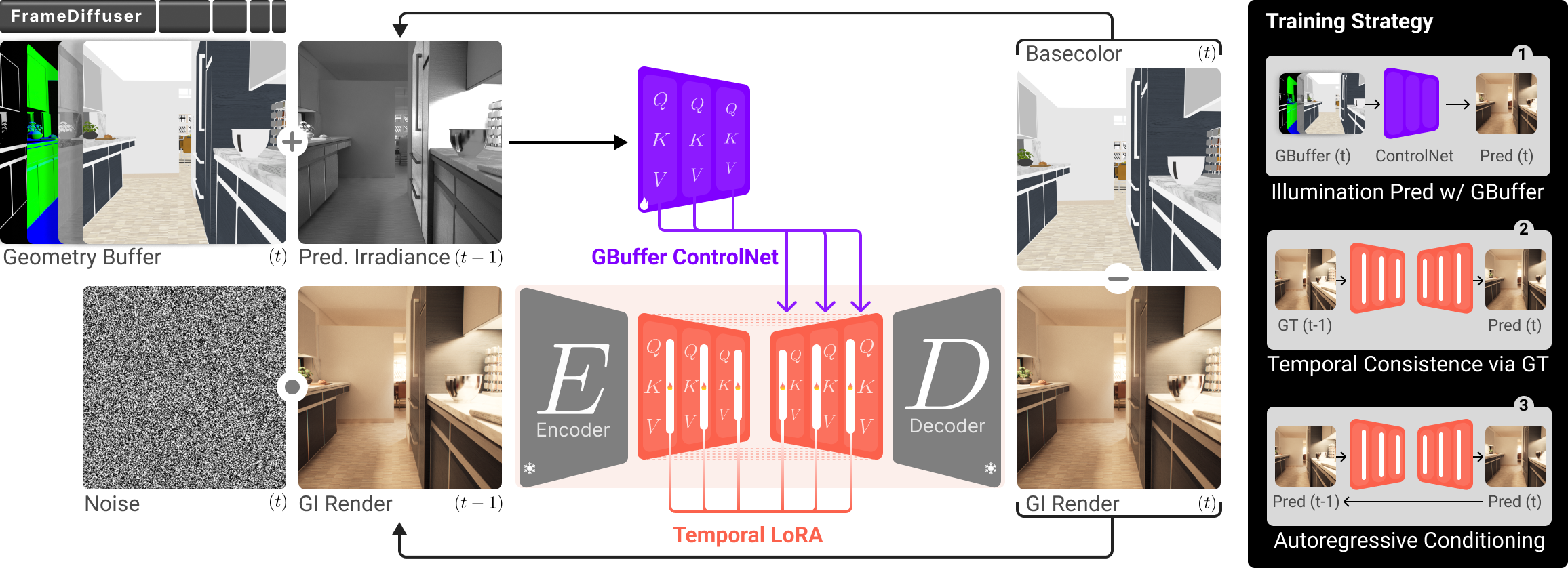}
    \caption{\textbf{FrameDiffuser Architecture} with dual conditioning: ControlNet processes 10-channel input comprising 9 G-buffer channels for structural guidance and 1 pred. irradiance channel for lighting guidance, computed from the previous frame's model output and basecolor. ControlLoRA conditions on the previous frame encoded in VAE latent space for temporal coherence. The generated output at time $t$ is used to compute the irradiance input for the next frame at time $t+1$, enabling autoregressive frame generation. The encoder $\mathcal{E}$ and decoder $\mathcal{D}$ represent the VAE components operating in latent space. The training strategy on the right shows our three-stage approach: first, we train ControlNet on the G-buffer to image translation task without irradiance. Second, we add ControlLoRA and irradiance for temporal conditioning. Third, we train autoregressively using the model's own generated frames as previous-frame inputs to make the model robust against its own generation errors.}
    \label{fig:architecture}
\end{figure*}

\section{Related Work}
\label{sec:related}

\noindent\textbf{Image Generation.}
Neural image generation has evolved dramatically from early GAN-based approaches to modern diffusion models. Generative Adversarial Networks (GANs)~\cite{goodfellow2014} pioneered high-quality neural image synthesis through adversarial training between generator and discriminator networks. However, GANs suffer from training instability and mode collapse, limiting their practical deployment.

The emergence of diffusion models~\cite{ho2020} marked a paradigm shift in generative modeling. These models learn to reverse a gradual noising process, achieving superior image quality and training stability compared to GANs~\cite{dhariwal2021}. Stable Diffusion~\cite{rombach2022} further improved efficiency by operating in a compressed latent space, enabling high-resolution generation with reduced computational requirements.

The success of diffusion models spurred research into fine-tuning techniques rather than training from scratch. ControlNet~\cite{zhang2023} introduced precise spatial control through additional conditioning networks, enabling structural guidance from depth maps, normal maps, and other inputs. LoRA~\cite{hu2022} provided efficient adaptation through low-rank matrices, reducing trainable parameters while maintaining performance.

\noindent\textbf{Video Generation.}
While image generation models produce individual frames with high quality, video generation requires temporal consistency across sequences. Video diffusion models like Stable Video Diffusion~\cite{blattmann2023} and earlier work by Ho et al.~\cite{ho2022video} generate entire sequences simultaneously, learning spatio-temporal patterns across complete videos. They produce complete sequences at once rather than frame-by-frame, making them unsuitable for interactive applications where future frames depend on user input.

\noindent\textbf{Autoregressive Generation for Interactive Applications.}
Interactive applications require user-driven frame-by-frame generation. Recent projects like Oasis AI~\cite{oasisai2024} and DeepMind's work~\cite{valevski2025} explore complete game world simulation but struggle with geometric consistency and control, suggesting that combining traditional rendering with AI augmentation may be more feasible.

A key challenge in autoregressive generation is preventing drift, where errors accumulate over time. Without mechanisms to handle distribution shift between training on perfect inputs and inference on generated inputs, model output degrades rapidly~\cite{zhang2025}.

\noindent\textbf{Self-Conditioning and Robustness Techniques.}
The autoregressive drift problem has motivated research into robustness techniques. Self Forcing~\cite{huang2025} and related work~\cite{chen2024neurips,denton2018,ardiffusion2025} demonstrate that training models on their own generated outputs bridges the train-test distribution gap, which is essential for stable autoregressive generation. Models without self-conditioning mechanisms fail at extended generation due to rapid error accumulation.

\noindent\textbf{G-Buffer-Based Neural Rendering.}
A parallel line of research explores using intermediate render passes from traditional graphics pipelines to guide neural generation. Deferred rendering~\cite{saito1990,shishkovtsov2005}, widely used in modern game engines~\cite{unreal_deferred,unity_deferred}, produces geometry buffers (G-buffer) that store per-pixel surface properties including depth, normals, basecolor, roughness, and metallicity.

EST-GAN~\cite{mittermueller2022} demonstrated that incorporating G-buffer data including depth, normal, and basecolor maps significantly improved visual quality compared to semantic maps alone. They pioneered environment-specific training on particular game environments to achieve superior consistency within those domains. However, this GAN-based approach was limited not only by a lack of temporal consistency but also by the inherent limitations of GANs compared to modern diffusion models.

Recent work has advanced G-buffer-based rendering using diffusion models. RGB$\leftrightarrow$X~\cite{zeng2024} presents a bidirectional framework handling both forward G-buffer-to-RGB and inverse RGB-to-G-buffer rendering. For the forward render task, they trained Stable Diffusion 2.1 with increased input channels to accommodate G-buffer data. Their intrinsic switch mechanism and channel dropout training enable generation with incomplete G-buffer sets, achieving high-quality single-image synthesis. However, RGB$\leftrightarrow$X focuses on individual images without temporal coherence. Xue et al.~\cite{xue2025} address these limitations through improved training strategies, notably demonstrating that using ControlNet for G-buffer conditioning achieves better results than direct channel concatenation.

DiffusionRenderer~\cite{liang2025} extends G-buffer-based neural rendering to video generation, processing normal, basecolor, depth, roughness, and metallic maps to generate temporally coherent sequences. Their lighting representation combines tone mapping, logarithmic intensity encoding, and directional information, achieving impressive visual quality for complete video sequences. However, like other video diffusion models, DiffusionRenderer does not operate autoregressively, making it unsuitable for interactive applications where future G-buffer depend on user input.

Summarizing, image models like RGB$\leftrightarrow$X generate high-quality individual images but lack temporal consistency. Video models like DiffusionRenderer excel at temporal coherence but require generating entire sequences at once. Our work fills this gap by combining G-buffer conditioning with autoregressive generation. Inspired by EST-GAN's environment-specific training~\cite{mittermueller2022}, we train specialized models for individual environments rather than attempting broad generalization.

\begin{figure*}[t]
    \centering
    \includegraphics[width=\textwidth]{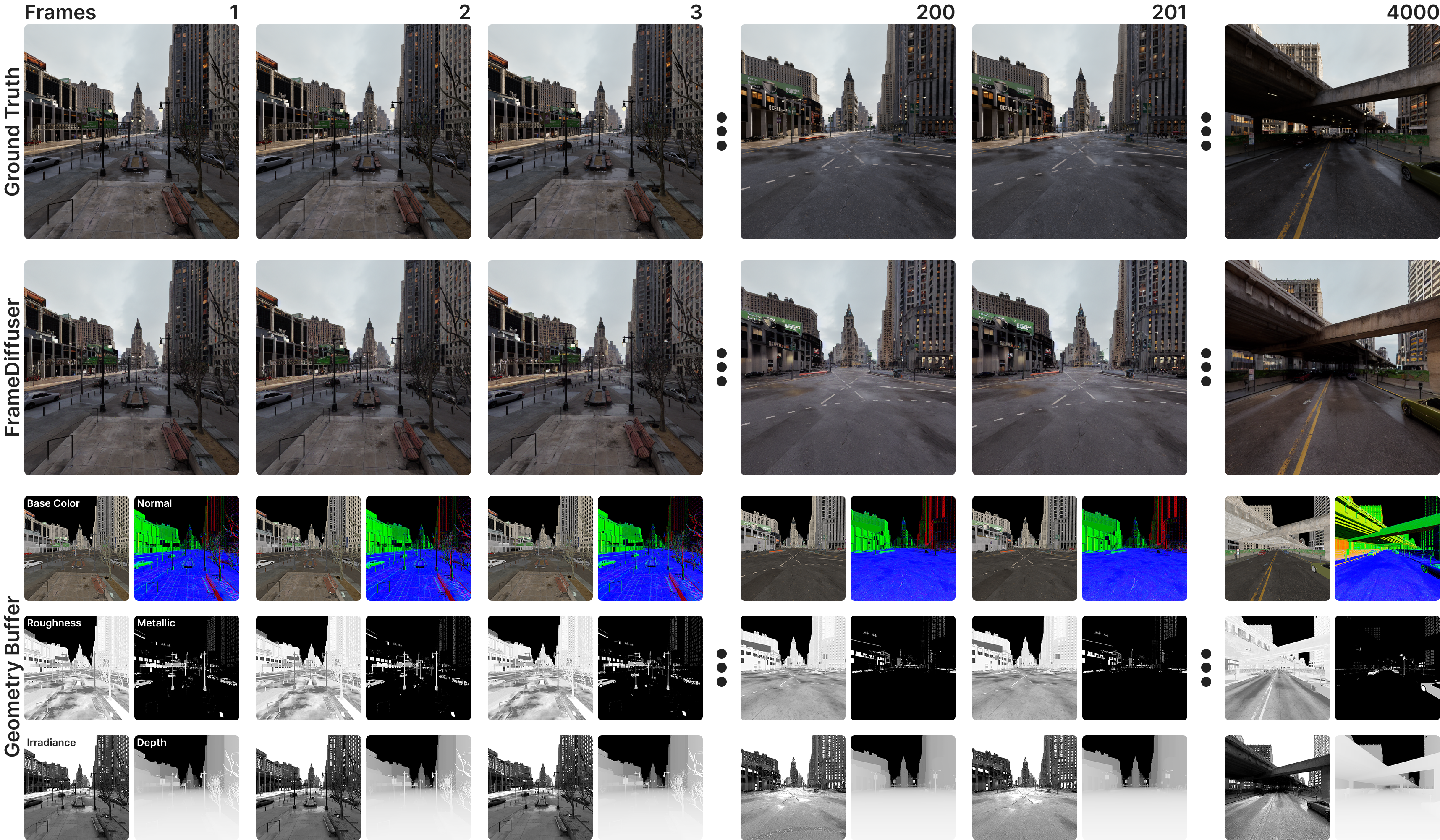}
    \caption{\textbf{Qualitative Results} showing FrameDiffuser's autoregressive generation across multiple frames, including long-term stability at frame 4000. From top to bottom: ground truth, our autoregressive output, and the G-buffer channels (basecolor, Normal, Depth, Roughness, Metallic) alongside computed Irradiance. The model maintains temporal consistency and accurate material properties across extended sequences.}
    \label{fig:qualitative}
\end{figure*}

\section{Method}
\label{sec:method}

We present FrameDiffuser, a neural rendering framework for autoregressive frame generation conditioned on G-buffer data. Our dual-conditioning architecture separates structural guidance from temporal coherence, enabling stable frame-by-frame generation for interactive applications. Figure~\ref{fig:architecture} illustrates the overall architecture.

\subsection{Architecture}

FrameDiffuser builds on Stable Diffusion 1.5, which is a pure text-to-image diffusion model, augmenting it with two complementary conditioning mechanisms that enable autoregressive frame generation.

\textbf{ControlNet for Structural Guidance.} We employ ControlNet~\cite{zhang2023} to process 10-channel input comprising basecolor, Normals, Depth, Roughness, Metallic, and an additional irradiance channel. The first nine channels represent common G-buffer data: basecolor (3 channels), Normals (3 channels), Depth, Roughness, and Metallic. ControlNet creates a trainable copy of the UNet encoder with zero-initialized connections, ensuring that training starts from the pre-trained state without disruption. This pathway establishes the mapping from geometric and material properties to photorealistic appearance.

\textbf{Irradiance as Temporal Lighting Condition.} We introduce an irradiance map $I_{t-1}$ that encodes temporal lighting information. Unlike RGB$\leftrightarrow$X~\cite{zeng2024} which can use ground-truth irradiance channels making lighting approximation for the forward render task almost trivial, we derive irradiance from the model's previous generated output and the basecolor channel, creating a lighting intensity map that provides ControlNet with strong guidance about shadow and specular locations:
\begin{equation}
I_{t-1}(x,y) = \frac{L(F_{t-1}(x,y))}{L(C_{t-1}(x,y)) + \epsilon},
\label{eq:irradiance}
\end{equation}
where $L(\cdot)$ converts RGB to grayscale, $F_{t-1}$ is the previous generated output, $C_{t-1}$ is the previous basecolor, and $\epsilon = 10^{-6}$. 
Values are clamped to [0, 2] and normalized for network input. This provides ControlNet with temporal lighting guidance while maintaining separation from ControlLoRA's role in temporal appearance consistency. 

\textbf{ControlLoRA for Temporal Coherence.} We use ControlLoRA~\cite{wu2024} to maintain temporal consistency by conditioning on the previous frame. Low-rank adaptation matrices applied to convolutional and linear layers throughout the UNet enable parameter-efficient fine-tuning. The previous frame is encoded through the VAE and concatenated to the noisy latent input in latent space, providing temporal appearance information and light color consistency at every denoising step. ControlLoRA nudges the generation to remain temporally consistent, particularly regarding lighting color and overall appearance.

Generation initiates from a single starting frame and its corresponding irradiance, from which all subsequent frames are synthesized autoregressively.

\subsection{Training Strategy}

We employ a three-stage training methodology where irradiance conditioning plays a central role in preventing temporal dependency and enabling stable autoregressive generation. 

\textbf{Stage 1: Structural Learning with Black Irradiance.} Only ControlNet trains while the base UNet remains frozen. ControlLoRA is not yet introduced. We provide ControlNet with the nine G-buffer channels and use all zeros for the irradiance channel, forcing ControlNet to learn the core translation from G-buffer geometry and material properties to photorealistic rendering. Starting with black irradiance ensures that ControlNet establishes strong reliance on the deterministic G-buffer guidance for the rendering task. This ensures the previous frame data serves only for temporal consistency while G-buffer remain the primary structural guidance.

\textbf{Stage 2: Temporal Coherence Introduction.} Both ControlNet and ControlLoRA train together. We now provide irradiance computed from previous frames to ControlNet's temporal channel. Thus, ControlNet receives the 9 G-buffer channels plus 1 irradiance channel. ControlLoRA receives the previous frame encoded in VAE latent space and concatenated to the noisy latent. By injecting noise to the previous frame, we force the model to still effectively utilize the G-buffer, as the irradiance and previous frame latent do not provide completely reliable data.

\textbf{Stage 3: Self-Conditioning for Robustness.} Both ControlNet and ControlLoRA continue training together. We now introduce self-conditioning by periodically injecting generated frames into the training that are produced using the current model weights. The previous frame latent serves as ControlLoRA conditioning input. Irradiance maps are computed from these generated frames and fed to ControlNet. The generated frames are now used recursively as conditioning for subsequent generated training samples. Training on the model's own imperfect outputs addresses the distribution mismatch between perfect training data and autoregressive deployment. The model becomes resilient to its own artifacts, preventing exponential quality degradation during extended autoregressive sequences.

This training strategy ensures ControlNet first learns the translation from G-buffer to photorealistic rendering, while ControlLoRA nudges the generation toward temporal consistency with the previous frame.

\begin{figure*}[ht]
    \centering
    \includegraphics[width=\textwidth]{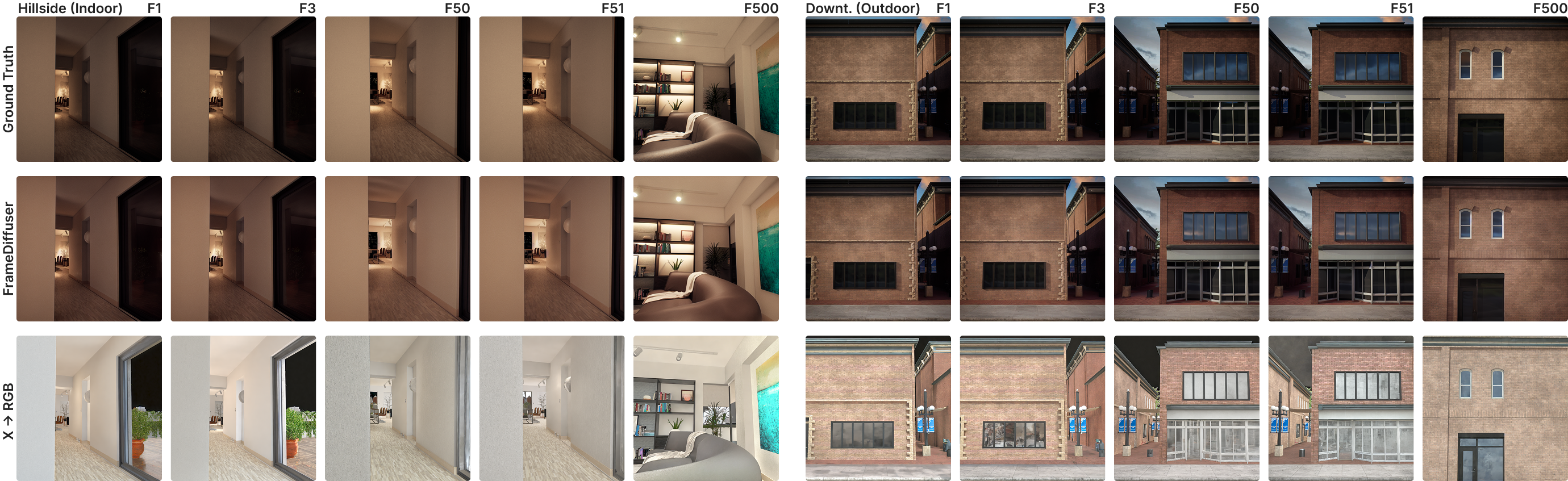}
    \caption{\textbf{Qualitative Comparison} with X→RGB across Downtown West (urban) and Hillside Sample (indoor) environments. Our method achieves high-detail lighting while maintaining temporal consistency across frames over long sequences, while X→RGB applies more uniform lighting. 
    }
    \label{fig:comparison}
\end{figure*}

\subsection{Temporal Coherence Mechanisms}

\textbf{Temporal Offset Sampling.} We sample temporal offsets $\delta \in \{-2, ..., 2\}$ with higher weights for adjacent frames to prevent overfitting to single-frame transitions.

\textbf{Sky Region Handling.} Sky regions present a challenge as deferred rendering provides no geometric data for skyboxes, resulting in empty G-buffer regions. Without geometric guidance, sky regions exhibit temporal inconsistencies. We address this through adaptive masking that injects sky information into the basecolor channel:
\begin{equation}
M_{ij} = \begin{cases}
1 & \text{if } \|\mathbf{c}_t^{(ij)}\|_{\infty} < \tau \text{ and } d_t^{(ij)} < \tau \\
0 & \text{otherwise}
\end{cases}
\end{equation}

where $\tau = 10/255$ identifies black pixels in basecolor $\mathbf{c}_t$ and Depth $d_t$. The modified basecolor becomes:

\begin{equation}
\tilde{\mathbf{c}}_t^{(ij)} = (1 - M_{ij}) \cdot \mathbf{c}_t^{(ij)} + M_{ij} \cdot \mathbf{x}_t^{(ij)}
\end{equation}

This provides ControlNet with complete scene information while maintaining clear differentiation through unchanged Depth, Normal, and material channels.

\textbf{Noise Injection.} We apply noise to the previous RGB frame before irradiance computation (Equation~\ref{eq:irradiance}):
\begin{equation}
\mathbf{x}'_{t-\delta} = \mathbf{x}_{t-\delta} + \sigma \cdot \boldsymbol{\epsilon}, \quad \sigma \sim \mathcal{U}(0.0, 0.2), \quad \boldsymbol{\epsilon} \sim \mathcal{N}(0, \mathbf{I})
\end{equation}

Both conditioning paths use the same noisy frame, encouraging stronger G-buffer utilization.

\section{Experiments and Results}
\label{sec:experiments}

We evaluate FrameDiffuser on autoregressive frame generation, comparing against baseline methods and analyzing our three-stage training strategy on temporal stability and generation quality.

\subsection{Experimental Setup}

\textbf{Datasets.} We train and evaluate separate models for each of our six Unreal Engine 5 environments: Electric Dreams~\cite{electricdreams2023}, City Sample~\cite{citysample2022}, Hillside Sample Project~\cite{hillsidesample}, Downtown West~\cite{downtownwest}, City Park~\cite{citypark}, and Derelict Corridor~\cite{derelictcorridor}. Each environment provides held-out validation splits covering diverse lighting conditions, material distributions, and geometric complexity.

\textbf{Metrics.} We employ standard image quality metrics: SSIM (Structural Similarity Index) for structural preservation, PSNR (Peak Signal-to-Noise Ratio) for pixel-level accuracy, and LPIPS (Learned Perceptual Image Patch Similarity) for perceptual alignment.

\textbf{Baselines.} We compare against X→RGB from RGB$\leftrightarrow$X~\cite{zeng2024}, a recent neural rendering method using image diffusion models for G-buffer-based synthesis.

\textbf{Implementation.} Training samples consist of previous frame and G-buffer pairs to generate the next frame, with samples shuffled during training. We build upon the pretrained Stable Diffusion 1.5~\cite{rombach2022} with ControlNet (10-channel input) and ControlLoRA (rank-64). Training uses batch size 2 with gradient accumulation of 4, AdamW optimizer (weight decay $10^{-2}$), and cosine learning rate scheduling. Stage 1 trains ControlNet only with black irradiance (40k steps, LR $2 \times 10^{-5}$). Stage 2 introduces temporal conditioning with real irradiance and noise injection (10k steps). Stage 3 adds self-conditioning with 50\% generated frames (30k steps). Training utilized NVIDIA RTX 4090 and A100 GPUs. Inference employs 10 denoising steps with DPMSolver, achieving approximately 1 frame per second on an RTX 4090.

\subsection{Quantitative Results}

\begin{table}[h]
\centering
\small
\caption{\textbf{Quantitative Comparison} on autoregressive frame generation. Results are averaged across validation sets from FrameDiffuser's training distribution.}
\label{tab:quantitative}
\begin{tabular}{lccc}
\toprule
\textbf{Method} & \textbf{PSNR ↑} & \textbf{SSIM ↑} & \textbf{LPIPS ↓} \\
\midrule
X→RGB~\cite{zeng2024} & 8.60 & 0.3566 & 0.5150 \\
FrameDiffuser & \textbf{18.34} & \textbf{0.6377} & \textbf{0.2129} \\
\bottomrule
\end{tabular}

\end{table}
Table~\ref{tab:quantitative} shows a quantitative comparison on held-out validation sets from each model's training environment. FrameDiffuser achieves higher SSIM and lower LPIPS compared to X→RGB, indicating better structural preservation and perceptual quality, demonstrating the effectiveness of our approach. 

\begin{figure}[t]
    \centering
    \includegraphics[width=\columnwidth]{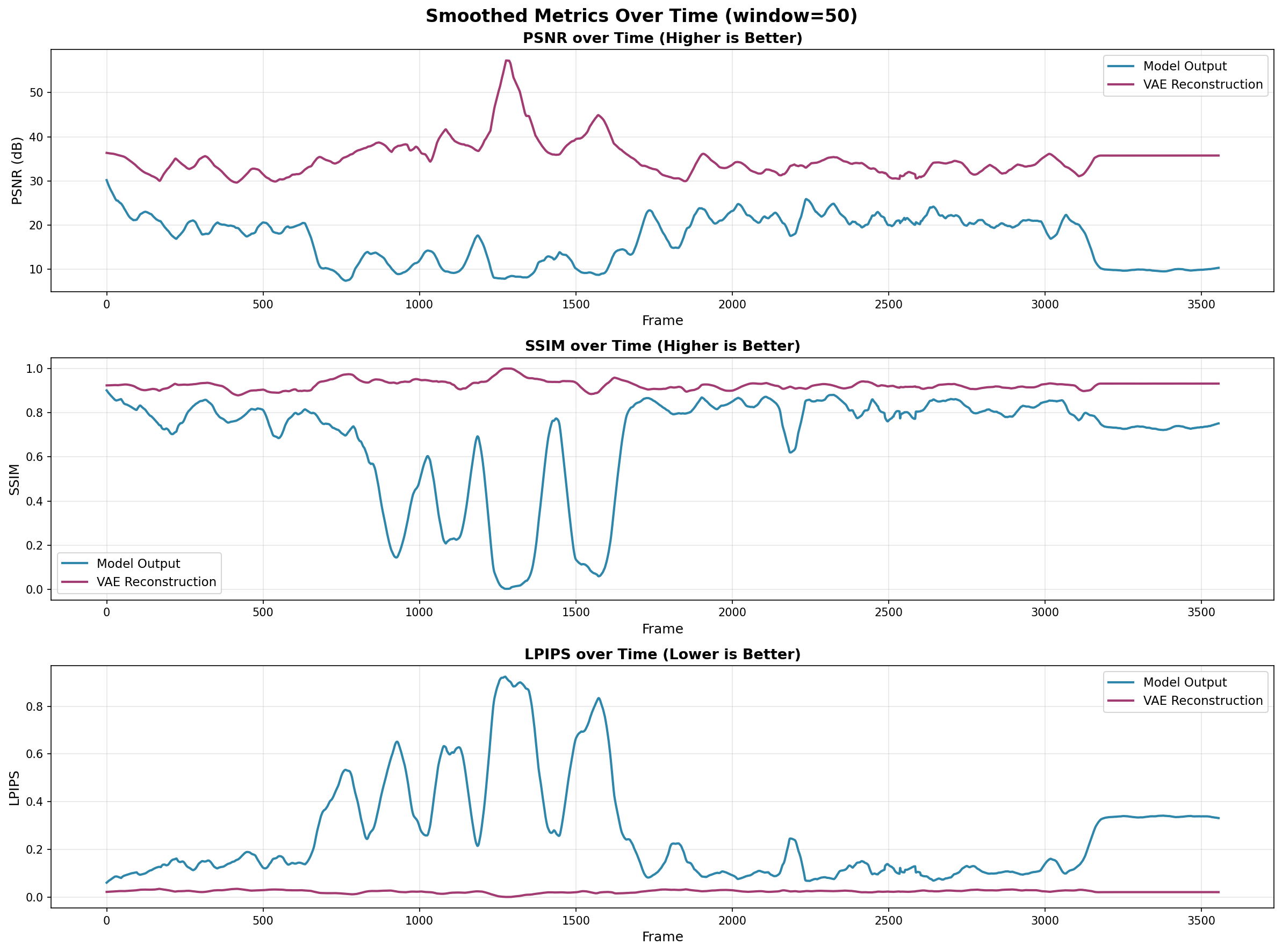}
\caption{\textbf{Temporal Stability Analysis} over 3000+ consecutive validation frames of pure autoregressive generation inside the Hillside Sample Project Environment~\cite{hillsidesample} Figure~\ref{fig:comparison}. We compare our model output against pure VAE reconstruction to measure encoder degradation. Metrics show degradation between frames 800--1700 when the camera enters very dark rooms; the model's bias towards lit areas causes it to insufficiently capture extreme darkness. After frame 3000, all metrics degrade when camera movement reduces and G-buffer changes become minimal, causing error accumulation.}
\end{figure}

\subsection{Ablation Studies}
\subsubsection{Self-Conditioning and Noise Injection}
We evaluate the impact of self-conditioning and noise injection on autoregressive generation quality using the Electric Dreams environment. Table~\ref{tab:selfcond_ablation} compares models trained with and without these mechanisms on the MidRange validation sequence.

\begin{figure}[t]
    \centering
    \includegraphics[width=\columnwidth]{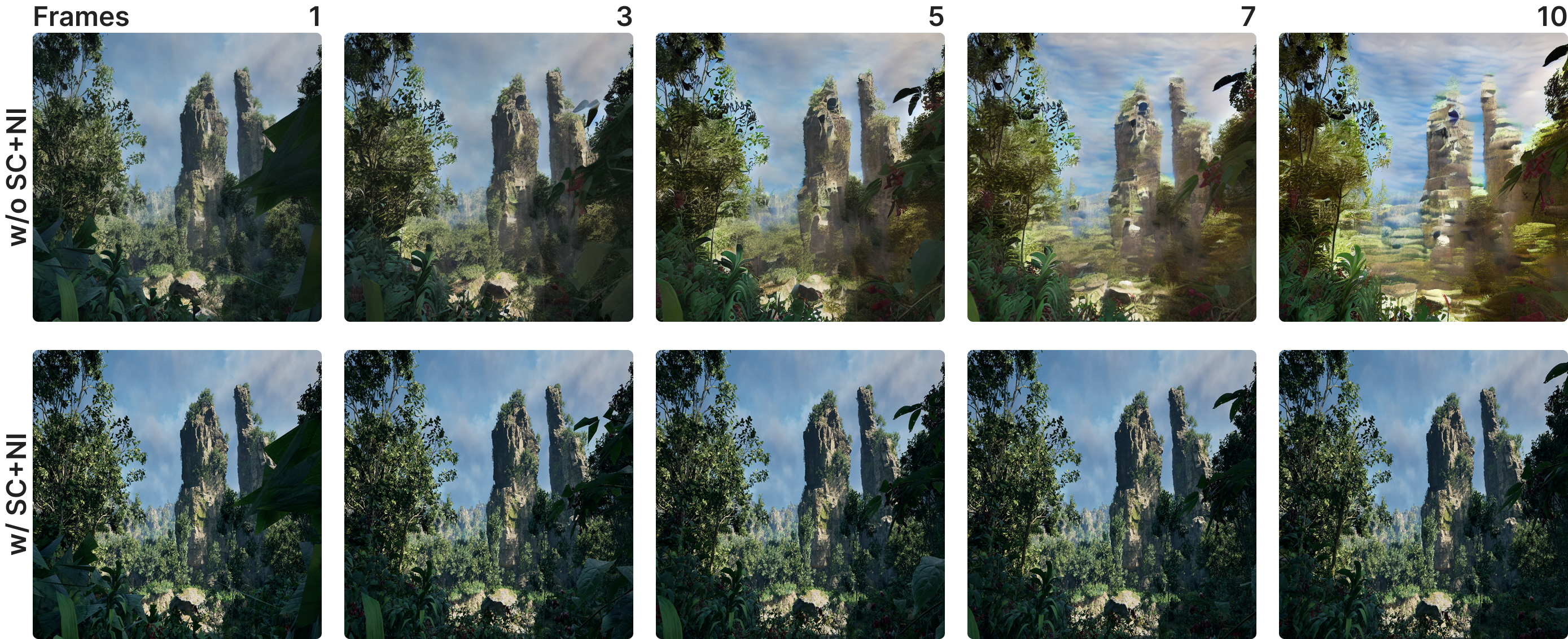}
    \caption{\textbf{Self-conditioning (SC) and Noise Injection (NI) Impact} on autoregressive generation. Without these mechanisms, severe degradation occurs within 5 frames. With self-conditioning and noise injection, quality remains stable over extended sequences.}
    \label{fig:selfcond_ablation}
\end{figure}

Without self-conditioning and noise injection, autoregressive generation exhibits rapid quality degradation within 5 frames, as the model trained exclusively on ground-truth inputs cannot process its own imperfect outputs during inference. With self-conditioning and noise injection incorporated during training phases 2 and 3, quality remains stable over extended sequences (Figure~\ref{fig:selfcond_ablation}). These mechanisms yield substantial improvements across all metrics: PSNR increases by 5.87 dB, SSIM improves by 40.6\%, and LPIPS decreases by 41.3\%, demonstrating that the model learns to handle imperfect inputs and maintain generation quality through temporal information propagation.

\begin{table}[h]
\centering
\small
\caption{\textbf{Self-conditioning and Noise Injection Ablation} on MidRange sequence (Electric Dreams~\cite{electricdreams2023}, 589 frames). Metrics averaged over autoregressive generation.}
\label{tab:selfcond_ablation}
\begin{tabular}{lccc}
\toprule
\textbf{Method} & \textbf{PSNR ↑} & \textbf{SSIM ↑} & \textbf{LPIPS ↓} \\
\midrule
Without SC + NI & 12.29 & 0.323 & 0.431 \\
With SC + NI & \textbf{18.16} & \textbf{0.454} & \textbf{0.253} \\
\midrule
Improvement & +47.8\% & +40.6\% & +41.3\% \\
\bottomrule
\end{tabular}
\end{table}

\subsubsection{Irradiance Conditioning}
We evaluate the contribution of our irradiance mechanism by comparing models trained with and without the irradiance channel. Table~\ref{tab:irradiance_ablation} shows results on 900 validation frames from the City Park~\cite{citypark} environment, chosen for its combination of dense vegetation and urban structures. The irradiance channel provides small but consistent improvements across all metrics, which also proves the strong guidance G-buffer provide alone.

\begin{table}[t]
\centering
\small
\caption{\textbf{Irradiance Conditioning Ablation} on City Park environment validation sequences with a total of 900 frames. Metrics averaged over autoregressive generation.}
\label{tab:irradiance_ablation}
\begin{tabular}{lccc}
\toprule
\textbf{Method} & \textbf{PSNR ↑} & \textbf{SSIM ↑} & \textbf{LPIPS ↓} \\
\midrule
Without Irradiance & 19.443 & 0.435 & 0.255 \\
With Irradiance & \textbf{19.998} & \textbf{0.440} & \textbf{0.246} \\
\bottomrule
\end{tabular}
\end{table}

\begin{figure*}[ht]
    \centering
    \includegraphics[width=\textwidth]{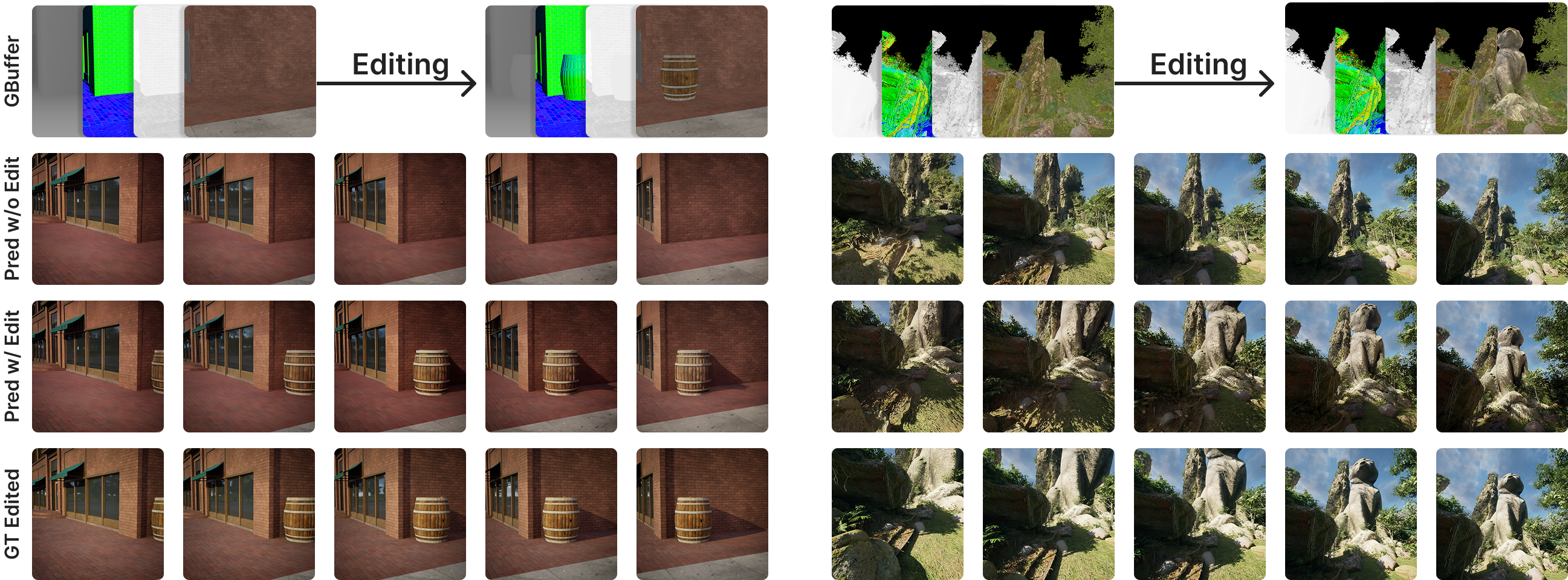}
    \caption{\textbf{Scene Editing} when objects are added to the scene through G-buffer modifications, FrameDiffuser automatically synthesizes appropriate lighting, shading, and cast shadows. Left: A barrel insertion demonstrates correct lighting direction and shadow casting. Right: A large concrete statue shows the model's ability to infer complex shadow patterns and lighting intensities on geometric additions, maintaining photorealistic consistency.}
    \label{fig:scene_edit}
\end{figure*}

\subsection{Qualitative Results}

Figure~\ref{fig:qualitative} shows FrameDiffuser outputs across multiple frames, including long-term generation at frame 4000. The model maintains temporal consistency and accurate geometric and material properties in extended sequences. Irradiance conditioning provides stable lighting cues, while G-buffer inputs ensure geometric consistency. FrameDiffuser can convincingly approximate realistic shadows as can be seen with the overpass in frame 4000 and understands to effectively use material properties, as can be observed with the effect of different roughness scales on the road and sidewalk and the shiny reflection of the car. This becomes more apparent when one looks at the irradiance channel, which is computed by the previous model output and the previous base color input. 

Beyond the geometric and material information in G-buffer, FrameDiffuser synthesizes photorealistic lighting including global illumination, shadows, reflections, and atmospheric effects that would traditionally require expensive ray tracing or pre-baked lightmaps.

\subsection{Scene Editing Capabilities}

Figure~\ref{fig:scene_edit} demonstrates FrameDiffuser's ability to handle interactive scene modifications through G-buffer editing. When objects are added to the scene, such as the barrel insertion shown on the left, the model automatically infers lighting direction to generate appropriate shading and cast shadows. The cat statue example on the right shows the model's capability to synthesize complex shadow patterns and lighting intensities on large geometric additions. This enables artists to maintain full control over scene composition while FrameDiffuser handles the computationally expensive lighting synthesis automatically.

\subsection{Comparison with Baselines}

We compare FrameDiffuser with X→RGB~\cite{zeng2024} in Figure~\ref{fig:comparison}. X→RGB produces images that appear artificially flat, closely resembling the raw G-buffer inputs with uniform lighting, lacking the rich lighting variation, shadow depth, and atmospheric effects present in photorealistic rendering. X→RGB also exhibits severe temporal inconsistencies: in the Hillside indoor example, lighting changes drastically between Frame 1 and Frame 3, with door frame shadows appearing and disappearing, and floor lighting patterns shifting completely. In contrast, FrameDiffuser generates natural-looking scenes with realistic global illumination while maintaining temporal consistency across frames.

\section{Conclusion}
\label{sec:conclusion}

FrameDiffuser enables frame-by-frame neural rendering for interactive applications through G-buffer conditioning and autoregressive generation. Our dual-conditioning architecture combines ControlNet for geometric guidance with ControlLoRA for temporal coherence, achieving stable generation where existing approaches fall short: single-image models like RGB$\leftrightarrow$X lack temporal consistency, while video models like DiffusionRenderer cannot accommodate user-driven frame generation. 

Our three-stage training strategy with self-conditioning proves essential for bridging the train-test distribution gap, enabling temporal consistency across extended sequences. We specialize models to individual environments rather than pursuing broad generalization, reflecting practical deployment needs where consistent rendering within specific visual domains matters more than moderate generalization across all environments.

\subsection{Limitations and Future Work}
Our environment-specific approach prioritizes consistency over generalization, requiring separate models for different visual styles. While current inference speeds reach approximately 1 frame per second on consumer hardware, distillation techniques and architectural optimizations present clear paths toward real-time performance. 

Future work could combine video diffusion architectures like 
FramePack~\cite{zhang2025} with our G-buffer conditioning 
approach, or explore integration with recent 3D scene generation and 
reconstruction methods~\cite{dihlmann2024signerf, engelhardt2025svim3d}, 
merging DiffusionRenderer's temporal modeling strengths with interactive 
generation requirements. Further, experimenting if world knowledge capabilities like changing scene style via prompting are still accessible are interesting directions to explore in future research. 

\begin{section}{Acknowledgements}
    \label{sec:acknowledgements}

    Funded by the Deutsche Forschungsgemeinschaft (DFG, German Research Foundation) under Germany's Excellence Strategy – EXC number 2064/1 – Project number 390727645.
    This work was supported by the German Research Foundation (DFG): SFB 1233, Robust Vision: Inference Principles and Neural Mechanisms, TP 02, project number: 276693517.
    This work was supported by the Tübingen AI Center.
    The authors thank the International Max Planck Research School for Intelligent Systems (IMPRS-IS) for supporting Jan-Niklas Dihlmann.
\end{section}

{
    \small
    \bibliographystyle{ieeenat_fullname}
    \bibliography{main}
}

\clearpage
\setcounter{page}{1}
\renewcommand\thesection{\Alph{section}}
\setcounter{section}{0}
\maketitlesupplementary

\begin{figure*}[h]
    \centering
    \includegraphics[width=0.90\textwidth]{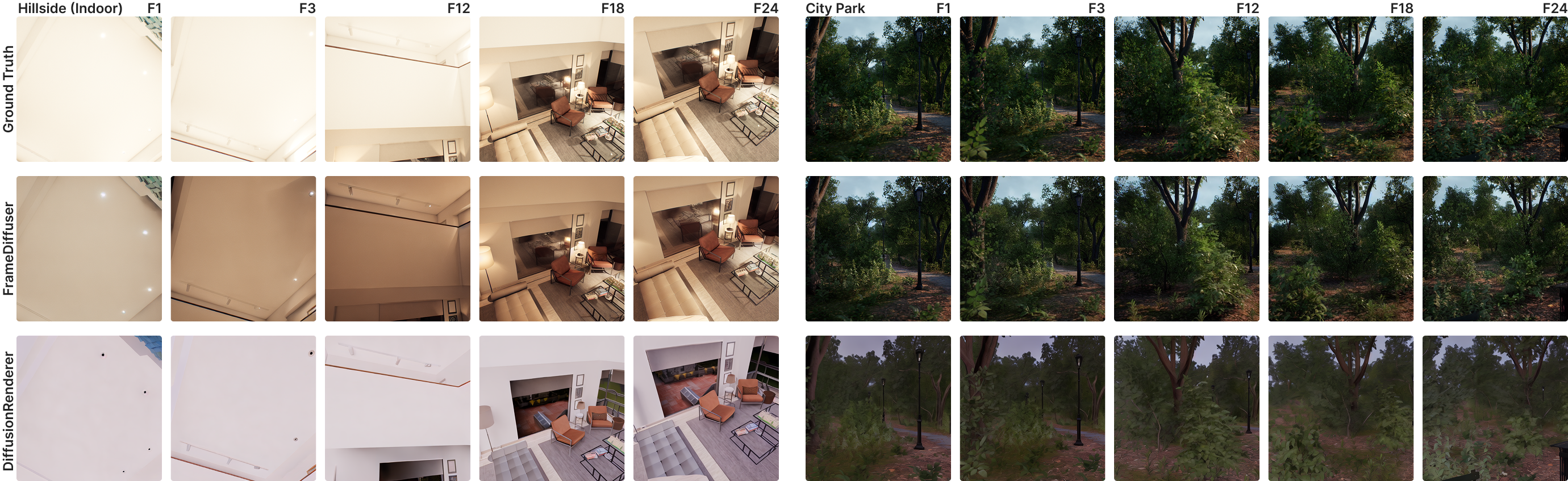}
    \caption{\textbf{DiffusionRenderer Comparison} across two environments. GT: Ground Truth, Ours: FrameDiffuser, DR: DiffusionRenderer~\cite{liang2025}. FrameDiffuser maintains closer alignment with ground truth lighting and atmospheric effects. See accompanying videos \texttt{DiffusionRenderer\_comparison\_CityPark.mp4} and \texttt{DiffusionRenderer\_comparison\_Hillside.mp4}.}
    \label{fig:diffusionrenderer_comparison}
\end{figure*}

This appendix offers supplementary information on FrameDiffuser. It provides a comparison with DiffusionRenderer~\cite{liang2025} explaining our methodological differences (Sec.~\ref{sec:diffusionrenderer}), presents out-of-distribution generalization results (Sec.~\ref{sec:ood}), analyzes temporal flicker and its sources (Sec.~\ref{sec:flicker_analysis}), presents GBufferDiffuser for inverse rendering (Sec.~\ref{sec:gbufferdiffuser}), explores style transfer capabilities (Sec.~\ref{sec:style_transfer}), and details our implementation specifications (Sec.~\ref{sec:implementation_details}).

\textbf{Additional Material} We wish to emphasize the inclusion of videos alongside this paper. Given the temporal nature of our results, these videos serve as the most effective medium for their evaluation. We strongly encourage viewing:
\begin{itemize}
    \item \texttt{FrameDiffuser\_hillside\_sequence.mp4} — Pure autoregressive generation in the Hillside Sample Project environment, showing ground truth versus our model versus basecolor for reference. Towards the end, the camera flies through a moderately dark room and then into a completely dark room. While the model approximates reasonable illumination, the predictions are far from the ground truth which is completely dark in these areas. Such mismatches can be observed in the Temporal Stability Analysis figure.
    \item \texttt{DiffusionRenderer\_comparison\_CityPark.mp4} — Comparison between FrameDiffuser and DiffusionRenderer on City Park environment sequences, demonstrating FrameDiffuser's superior lighting accuracy.
    \item \texttt{DiffusionRenderer\_comparison\_Hillside.mp4} — Comparison between FrameDiffuser and DiffusionRenderer on Hillside Sample Project sequences.
    \item \texttt{vae\_flicker\_citysample.mp4} — Video showing the temporal flicker artifacts in the City Sample metropolitan environment. The video compares our model output against pure VAE reconstruction, demonstrating that the VAE encoder-decoder cycle is the primary source of frame-to-frame inconsistencies, particularly in scenes with high spatial frequencies.
\end{itemize}

\section{DiffusionRenderer Comparison}
\label{sec:diffusionrenderer}

We did not include DiffusionRenderer~\cite{liang2025} as a baseline in the main paper due to fundamental differences in generation paradigm. DiffusionRenderer is designed for video generation, processing complete sequences at once with access to all frames during generation. In contrast, FrameDiffuser operates autoregressively, generating each frame based only on the current G-buffer and the previously generated frame. This autoregressive approach is essential for interactive applications such as video games, where future frames cannot be predetermined before user input. The user's actions determine the next G-buffer state, making it impossible to provide future frame information during generation.

Despite these methodological differences, we conducted a quantitative comparison on 24-frame sequences, the maximum length DiffusionRenderer supports, even though FrameDiffuser can generate arbitrarily long sequences autoregressively. Results are averaged across validation sets from six environments.

\begin{table}[h]
\centering
\small
\caption{\textbf{DiffusionRenderer Comparison} on 24-frame sequences. Results averaged across validation sets from all six training environments.}
\label{tab:diffusionrenderer}
\begin{tabular}{lccc}
\toprule
\textbf{Method} & \textbf{PSNR $\uparrow$} & \textbf{SSIM $\uparrow$} & \textbf{LPIPS $\downarrow$} \\
\midrule
DiffusionRenderer~\cite{liang2025} & 13.03 & 0.4683 & 0.4583 \\
FrameDiffuser (Ours) & \textbf{20.96} & \textbf{0.6378} & \textbf{0.2030} \\
\bottomrule
\end{tabular}
\end{table}

Figure~\ref{fig:diffusionrenderer_comparison} shows qualitative comparisons across two sequences from City Park~\cite{citypark} and Hillside Sample Project~\cite{hillsidesample}. FrameDiffuser maintains closer alignment with ground truth lighting and atmospheric effects throughout the sequences, while DiffusionRenderer produces results with inconsistent lighting and reduced detail.

\section{Out-of-Distribution Generalization}
\label{sec:ood}
\begin{figure}[!htbp]
    \includegraphics[width=0.99\columnwidth]{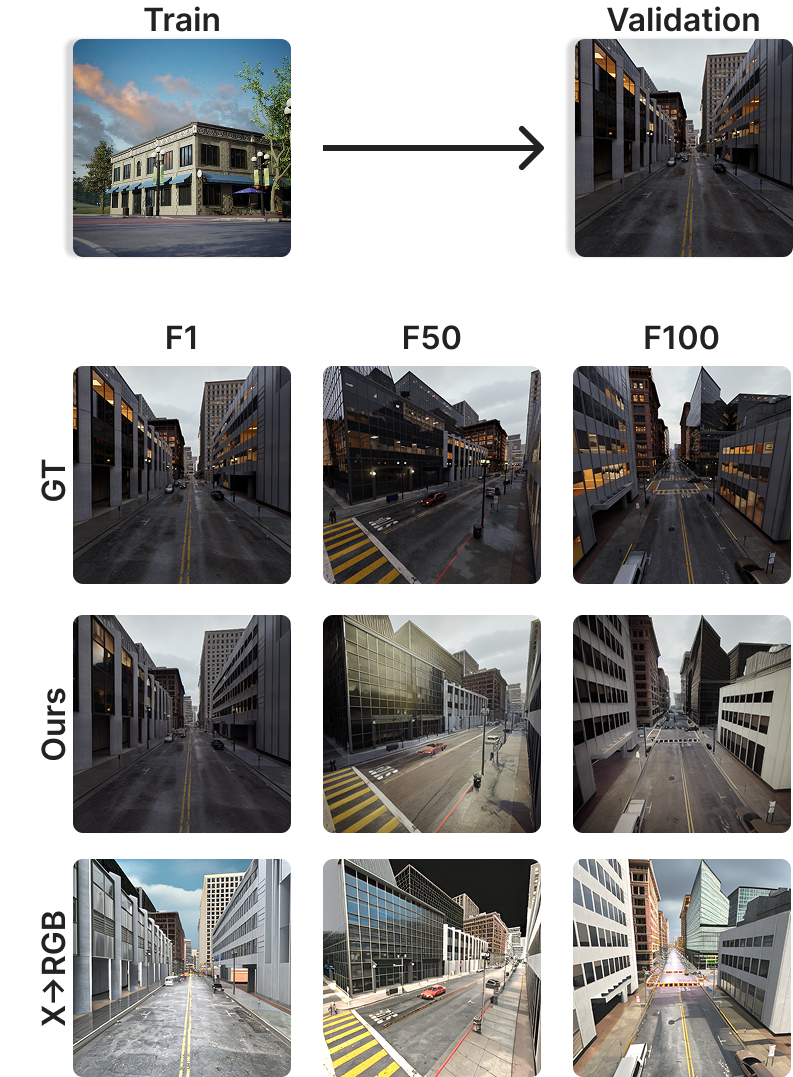}
    \caption{\textbf{Out-of-Distribution Generalization.} Top left: Downtown West training environment. Top right: Starting frame for generation. Bottom: Frames 1, 50, and 100 from a 100-frame autoregressive sequence on City Sample, showing GT (top row), FrameDiffuser (middle), and X$\rightarrow$RGB (bottom). The warm Downtown West style is slightly transferred to the cold City Sample environment.}
    \label{fig:out_of_distribution}
\end{figure}
We evaluate out-of-distribution generalization using a model trained on Downtown West~\cite{downtownwest} and tested on City Sample~\cite{citysample2022}. Downtown West features Pearl Street Mall-inspired architecture with warm, sunny lighting, trees, and colorful storefronts, while City Sample presents grey, cloudy skies and concrete-dominated metropolitan architecture. 

Figure~\ref{fig:out_of_distribution} shows the out-of-distribution generalization test. We use the prompt ``A photorealistic metropolitan city with skyscrapers and streets under overcast grey skies'' for generation. The figure displays an example frame from the Downtown West training environment (top left) and the starting frame for generation (top right). Below, we show frames 1, 50, and 100 comparing ground truth, our model, and X$\rightarrow$RGB. At frame 1, the model output remains close to the ground truth as it can derive the appropriate style from the starting frame. However, as the autoregressive generation progresses, the model increasingly diverges from the cold, grey City Sample aesthetic towards more of the warm, vibrant style learned from Downtown West. By frames 50 and 100, we observe the model introducing warmer lights and more saturated colors, demonstrating how the learned environment-specific characteristics gradually dominate the output during extended autoregressive generation.

Table~\ref{tab:ood_results} shows FrameDiffuser achieves superior performance despite training on a single and very different environment.

\begin{table}[!htbp]
\centering
\small
\caption{\textbf{Out-of-Distribution Performance.} Downtown West model evaluated on City Sample (100 frames).}
\label{tab:ood_results}
\begin{tabular}{lccc}
\toprule
\textbf{Method} & \textbf{PSNR $\uparrow$} & \textbf{SSIM $\uparrow$} & \textbf{LPIPS $\downarrow$} \\
\midrule
X$\rightarrow$RGB~\cite{zeng2024} & 7.52 & 0.255 & 0.442 \\
FrameDiffuser (Ours) & \textbf{14.17} & \textbf{0.454} & \textbf{0.363} \\
\bottomrule
\end{tabular}
\end{table}

\section{Temporal Flicker Analysis}
\label{sec:flicker_analysis}

We provide \texttt{vae\_flicker\_citysample.mp4} to demonstrate temporal flicker in our generated sequences. As can be seen when comparing to the pure VAE reconstruction, the VAE of Stable Diffusion 1.5~\cite{rombach2022} introduces flicker, especially in scenes with high spatial frequencies like the metropolitan city example in the video.
We strongly believe that with a different VAE this flicker could be substantially 
reduced or eliminated. However, training or fine-tuning a specialized VAE requires 
significant computational resources and falls outside the scope of this work, which 
focuses on the frame generation pipeline itself. Potential solutions include 
finetuned VAE decoders, such as the approach by Valevski~\emph{et al.}~\cite{valevski2025} for GameNGen, or VAEs that incorporate previous frame context. %

\section{GBufferDiffuser: Inverse Rendering}
\label{sec:gbufferdiffuser}

While forward rendering (G-buffer to RGB) is the main focus of this work, we also developed GBufferDiffuser for inverse rendering (RGB to G-buffer), similar to the inverse capabilities demonstrated by RGB$\leftrightarrow$X~\cite{zeng2024}. This system employs five independent ControlLoRA models with rank-128 matrices, each specialized for reconstructing a specific G-buffer component: basecolor, Depth, Normals, Roughness, and Metallic. Each model conditions on the final rendered image to reconstruct its respective component.

In contrast to generalist approaches that train a single model across multiple G-buffer types and environments, we train smaller specialized adapters for each component on a single environment. We trained each model for only 10k steps on the Hillside Sample Project~\cite{hillsidesample} environment with the same batch size and gradient accumulation as FrameDiffuser. This specialization approach allows for faster training while achieving superior results within the target domain, at the cost of generalization to arbitrary environments.

Table~\ref{tab:gbufferdiffuser} shows quantitative results comparing GBufferDiffuser against RGB$\rightarrow$X from RGB$\leftrightarrow$X~\cite{zeng2024}. GBufferDiffuser substantially outperforms the generalist baseline across all G-buffer components. Depth reconstruction is not supported by RGB$\leftrightarrow$X.

\begin{table}[!htbp]
\centering
\small
\caption{\textbf{GBufferDiffuser: Inverse Rendering} on Hillside Sample Project validation sequence. RGB$\rightarrow$X does not support Depth reconstruction.}
\label{tab:gbufferdiffuser}
\begin{tabular}{llccc}
\toprule
\textbf{Component} & \textbf{Method} & \textbf{PSNR $\uparrow$} & \textbf{SSIM $\uparrow$} & \textbf{LPIPS $\downarrow$} \\
\midrule
BaseColor & RGB$\rightarrow$X & 12.60 & 0.756 & 0.398 \\
& Ours & \textbf{19.39} & \textbf{0.792} & \textbf{0.172} \\
\midrule
Normals & RGB$\rightarrow$X & 5.01 & 0.146 & 0.700 \\
& Ours & \textbf{10.77} & \textbf{0.426} & \textbf{0.387} \\
\midrule
Roughness & RGB$\rightarrow$X & 13.32 & 0.679 & 0.527 \\
& Ours & \textbf{22.30} & \textbf{0.806} & \textbf{0.202} \\
\midrule
Metallic & RGB$\rightarrow$X & 6.42 & 0.017 & 0.917 \\
& Ours & \textbf{14.59} & \textbf{0.242} & \textbf{0.272} \\
\midrule
Depth & Ours & \textbf{12.92} & \textbf{0.772} & \textbf{0.496} \\
\bottomrule
\end{tabular}
\end{table}

\begin{figure}[!htbp]
    \centering
    \includegraphics[width=\columnwidth]{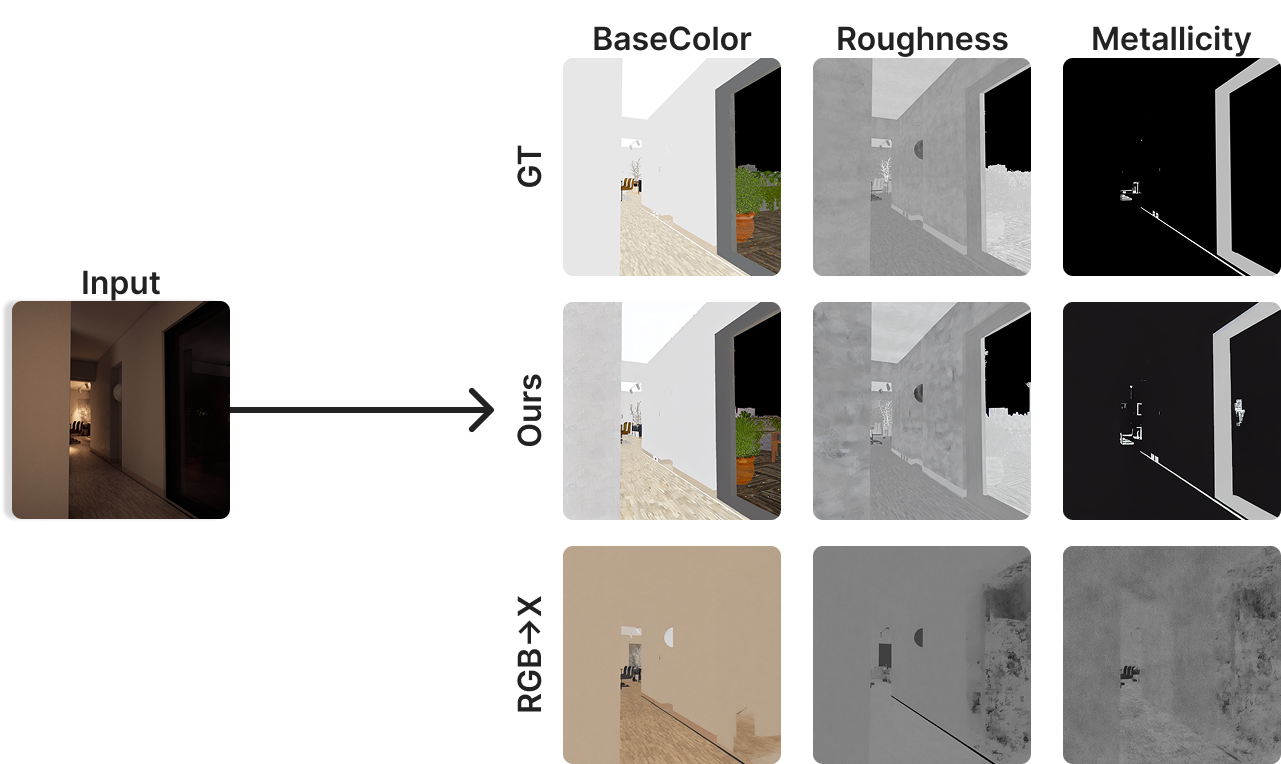}
    \caption{\textbf{GBufferDiffuser Qualitative Comparison.} GBufferDiffuser produces more accurate reconstructions with better shape and color prediction compared to RGB$\rightarrow$X~\cite{zeng2024}, due to specialization to the G-buffer type and environment.}
    \label{fig:gbufferdiffuser}
\end{figure}

Figure~\ref{fig:gbufferdiffuser} shows qualitative comparisons for basecolor, Roughness and Metallicity. GBufferDiffuser produces more accurate reconstructions with better preservation of fine details, while RGB$\rightarrow$X struggles particularly with Metallic reconstruction.

\section{Style Transfer}
\label{sec:style_transfer}

We explored artistic control through style transfer by applying first-frame augmentation. The process works as follows: given an original rendered frame, we apply Stable Diffusion's image-to-image transformation with a style-specific text prompt (e.g., ``Forest fire, flames, smoke, burning trees'') to create a stylized version. This stylized frame then serves as the previous frame input for the first generated frame, and generation continues autoregressively. Additionally, a fitting prompt for the new style is used.

\begin{figure}[!htbp]
    \centering
    \includegraphics[width=\columnwidth]{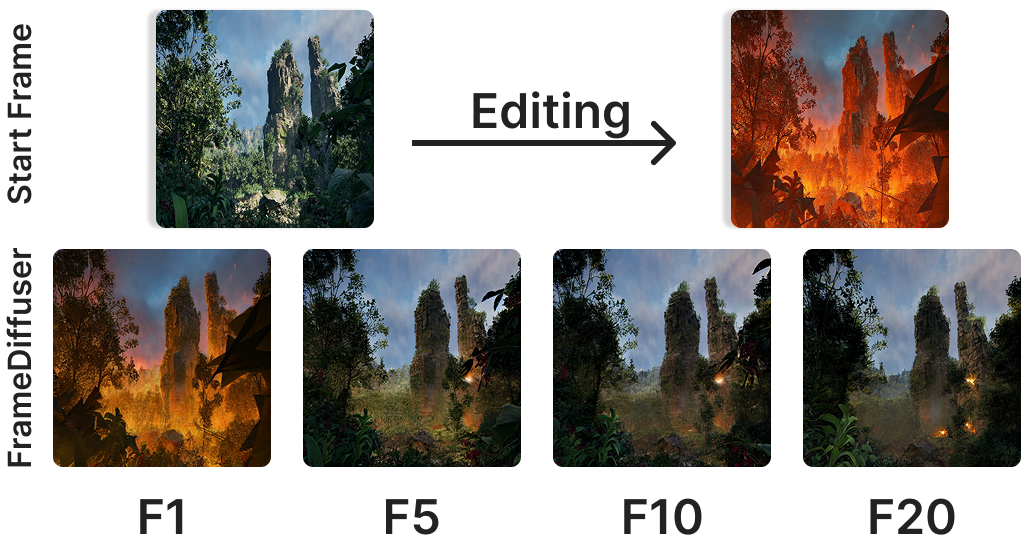}
    \caption{\textbf{Style Transfer: Burning Style.} The original frame is stylized via image-to-image transformation (Edited), then used as previous frame input for autoregressive generation. Fire effects persist through temporal propagation but gradually diminish as the model reverts to its training distribution.}
    \label{fig:style_transfer}
\end{figure}

Figure~\ref{fig:style_transfer} shows an example with a burning style. The original frame is transformed into a burning scene, which then conditions the subsequent autoregressive generation. Fire effects remain visible but are gradually diminishing as the model reverts to its training distribution.

The experiments revealed limited style transfer capabilities due to the model's specialization on training environments with fixed prompts. After training on a single prompt-environment combination, the model's ability to respond to alternative text conditioning decreased.

\section{Implementation Details}
\label{sec:implementation_details}

\textbf{Dataset Details} Table~\ref{tab:dataset} shows the training and validation set sizes for each environment. All sets consist of consecutive frame pairs with corresponding G-buffers. Validation sequences vary in length from 24 frames for short-term evaluation to over 3500 frames for long-term autoregressive stability analysis.

\begin{table}[h]
\centering
\small
\caption{\textbf{Dataset Statistics} for each training environment (rounded values).}
\label{tab:dataset}
\begin{tabular}{lcc}
\toprule
\textbf{Environment} & \textbf{Train} & \textbf{Validation} \\
\midrule
Electric Dreams~\cite{electricdreams2023} & 40,000 & 600 \\
City Sample~\cite{citysample2022} & 46,000 & 11,500 \\
City Park~\cite{citypark} & 23,000 & 5,800 \\
Derelict Corridor~\cite{derelictcorridor} & 32,000 & 8,100 \\
Hillside Sample Project~\cite{hillsidesample} & 18,100 & 3,500 \\
Downtown West~\cite{downtownwest} & 41,000 & 10,000 \\
\bottomrule
\end{tabular}
\end{table}

\textbf{Training Configuration} The training configuration was optimized for NVIDIA RTX 4090 and A100 GPUs. We use batch size 2 with gradient accumulation of 4 for an effective batch size of 8. The AdamW optimizer is configured with weight decay $10^{-2}$ and cosine learning rate scheduling. Stage 1 trains ControlNet with learning rate $2 \times 10^{-5}$ for 40k steps. Stages 2 and 3 continue with reduced learning rates for 10k and 30k steps respectively.

The ControlLoRA configuration uses rank-64 matrices targeting all linear and convolutional layers. We employ gradient clipping with max\_norm = 1.0 and enable gradient checkpointing for memory optimization. Inference uses classifier-free guidance scale of 1.0, which achieved best results for in-distribution environments.

\textbf{Hardware and Performance} Training requires approximately 23GB VRAM when training both ControlNet and ControlLoRA components. Inference achieves approximately 1 frame per second on an RTX 4090 with 10 denoising steps using the DPMSolver scheduler.

\textbf{Optimization Potential} Significant optimization potential exists through methods not explored in this work, ranging from simple techniques like batching and mixed precision to more advanced approaches such as StreamDiffusion~\cite{kodaira2023streamdiffusion}, model distillation, or recent NVIDIA DLSS advancements. %

\end{document}